\documentclass[conference]{IEEEtran}
\IEEEoverridecommandlockouts

\usepackage{cite}
\usepackage{amsmath,amssymb,amsfonts}
\usepackage{algorithmic}
\usepackage{bm}
\usepackage{mathtools}
\DeclarePairedDelimiterX{\infdivx}[2]{(}{)}{%
  #1\;\delimsize|\delimsize|\;#2%
}
\usepackage{multirow}
\usepackage{breqn}
\usepackage{siunitx}

\usepackage[font=small,labelfont=bf]{caption}
\usepackage{subfig}

\usepackage{color,soul}

\usepackage{xcolor}
\def\BibTeX{{\rm B\kern-.05em{\sc i\kern-.025em b}\kern-.08em
    T\kern-.1667em\lower.7ex\hbox{E}\kern-.125emX}}
	
\usepackage{url}
	
\begin{document}

\title{Knowing when we do not know: Bayesian continual learning for sensing-based analysis tasks}

\author{\IEEEauthorblockN{Sandra Servia-Rodriguez\thanks{Work done while Sandra was a postdoctoral researcher at the Department of Computer Science and Technology of the University of Cambridge.}}
\IEEEauthorblockA{\textit{Dept. of Comp. Sci. \& Tech.} \\
\textit{University of Cambridge, UK} \\
sandra.servia-rodriguez@cl.cam.ac.uk}
\and
\IEEEauthorblockN{Cecilia Mascolo}
\IEEEauthorblockA{\textit{Dept. of Comp. Sci. \& Tech.} \\
\textit{University of Cambridge, UK} \\
cecilia.mascolo@cl.cam.ac.uk}
\and
\IEEEauthorblockN{Young D. Kwon}
\IEEEauthorblockA{\textit{Dept. of Comp. Sci. \& Tech.} \\
\textit{University of Cambridge, UK} \\
ydk21@cam.ac.uk}
}

\maketitle

\begin{abstract}
	
	Despite much research targeted at enabling conventional machine learning models to continually learn tasks and data distributions sequentially without forgetting the knowledge acquired, little effort has been devoted to account for more realistic situations where learning some tasks accurately might be more critical than forgetting previous ones. In this paper we propose a Bayesian inference based framework to continually learn a set of real-world, sensing-based analysis tasks that can be tuned to \emph{prioritize} the remembering of previously learned tasks or the learning of new ones. Our experiments prove the robustness and reliability of the learned models to adapt to the changing sensing environment, and show the suitability of using uncertainty of the predictions to assess their reliability.
		
\end{abstract}

\begin{IEEEkeywords}
Continual Learning, uncertainty, sensing-based applications
\end{IEEEkeywords}

\section{Introduction}
\label{sec:intro}

	\emph{Continual learning} enables machine learning models to learn tasks and data distributions sequentially, while being able to reuse and retain the knowledge acquired over time. This is important in many real-world situations where data become incrementally available over time, when data come from non-stationary distributions or when new related tasks need to be learned. Examples include sensing-based applications, i.e., applications that combine sensory type data and machine learning to learn precise models of human behaviour, as they often need to adapt to the continuously changing environment. A significant amount of research has been devoted to investigate how to prevent conventional ML models from forgetting the knowledge of previous tasks while learning new ones --a phenomenon known as \emph{catastrophic forgetting}. However, little effort has been put in investigating the feasibility of these models in real world settings, such as sensing domains.
	
	In real-world applications some inference tasks are more critical than others. For example, in some audio-based applications, higher accuracy might be desirable on recognising speakers than on detecting speech. Thus, if these tasks were to be learned sequentially, it might be useful to be able to tell the model to prioritize the learning of speaker recognition over speech detection. However, previous research has mainly focused on preventing models from forgetting the knowledge of previous tasks without accounting for situations where learning some tasks more accurately is preferable.
	
	Moreover, despite its groundbreaking achievements, most deep learning models do not represent uncertainty. Although this might seem less relevant in tasks such as distinguishing between cats and dogs, this is of key importance in more sensitive tasks such as those involving healthcare applications or autonomous vehicles. Bayesian models in continual learning have the natural ability to retain knowledge from previous tasks by treating it as the prior when learning the current task. More importantly, apart from inferring a given class within a task, Bayesian models also output the uncertainty of the prediction, making them especially useful to model sensitive tasks.

	In this work, we propose and evaluate a continual learning framework for sensing-based analysis tasks. We opt for a Bayesian inference based approach to learning that provides the model with the ability to tell when it does not know. More specifically, we adapt the Variational Continual Learning model proposed by Nguyen \emph{et al.}~\cite{nguyen2018variational}, which combines streaming variational inference~\cite{broderick2013streaming} and Monte Carlo variational inference for neural networks~\cite{blundell2015weight}, for sensing-based analysis tasks (\S\ref{sec:methodology}). We extend this work by increasing the flexibility of the model to prioritize the learning of new tasks or the remembering of previously learned ones. We do so by adding an extra hyperparameter to the loss function that \emph{regulates} the influence of the prior and the likelihood of the new data in the approximation of the posterior.

	We make the following contributions:

	\begin{itemize}
		\item  A framework for continual learning sensing-based analysis tasks that extends the Variational Continual Learning model in~\cite{nguyen2018variational} by adding an extra hyperparameter $\beta$ that weights the contribution of the likelihood and the KL-divergence term in the loss function. This 
		 allows to trade catastrophic forgetting by intransigence (ability to learn new tasks).
	 
		\item  An evaluation of our framework applied to sensed audio and activity data, respectively (\S\ref{sec:expSetup} and \S\ref{sec:evaluation}). Using several datasets of audio and activity, we incrementally learn a model to infer different audio tasks namely speaker identification, emotion recognition, stress detection and ambient scene analysis, and another to infer the activity-related tasks of human activity recognition and person identification. We show the effect of $\beta$ in the forget and intransigence of the models and tune this hyperparameter to maximize accuracy, and minimize forget and intransigence. We prove the robustness of our results by experimenting with different order of tasks to learn.  
	
		\item An exploratory analysis of the uncertainty of our predictions in which we compute, for each sample in the test set, the predictive entropy and the mutual information between the prediction and the posterior over the model parameters. Our results show that the uncertainty tends to be higher for those samples wrongly classified than for those correctly inferred. Thus, uncertainty can be used to accept or discard inferences.
 
		\item An study of the feasibility of deploying a trained continual learning model for audio tasks in an Android smartphone.
	
	\end{itemize}

	Our results will aid designers and developers of machine learning based applications that use sensing data in mobile and edge computing setups, especially for those where accurately assessing the uncertainty of the prediction is crucial.
	
\section{Related Work}
\label{sec:related}

This section covers relevant related work, which including advances in the application of deep learning to mobile sensor data, continual learning and uncertainty estimation.

\textbf{Ubiquitous Computing and Deep Learning.} 
Recent advances on computer vision and language processing using deep learning have motivated the application of this technology to mobile and sensor data. Examples include audio sensing tasks such as speaker recognition or emotion detection from speech~\cite{georgiev2014dsp}, human activity recognition from accelerometer data~\cite{wang2019deep}, or learning sleep stages from radio signals~\cite{zhao2017learning}. Apart from improving performance on existing tasks and enabling new ones, researchers have also worked on model and resource optimizations to run costly deep learning algorithms in resource-constrained devices~\cite{georgiev2017low, yao2018fastdeepiot}. However, most of the efforts have focused on optimizing inference, while little attention has been given to efficient on-device training to adapt models to changes in the distribution of the incoming data, a common issue in mobile applications with continuously changing context.

\textbf{Continual Learning for Deep Neural Networks.}
Continual or lifelong learning allows models to learn continually from a stream of data, being able to build on what was learned and without forgetting previously seen tasks. Conventional models tend to forget previous tasks while learning new tasks, a phenomenon known as \emph{catastrophic forgetting}. Solutions proposed to prevent catastrophic forgetting include (i) regularizing the loss, i.e., adjusting the learning rates to minimize changes in the parameters that are important for previous tasks~\cite{kirkpatrick2017overcoming,zenke2017continual,chaudhry2018riemannian,nguyen2018variational}, and (ii) storing and replaying examples from previous tasks (or artificially generating examples from these tasks) in order to maintain predictions invariant by means of distillation~\cite{rebuffi2017icarl,lopez2017gradient, shin2017continual}. However, most of these works focus on minimising gradual forgetting without accounting for situations where the learning of new tasks needs to be prioritised. We use a regularized-based method in our framework, in which we extend the variational continual learning framework of Nguyen \emph{et al.}~\cite{nguyen2018variational} to allow for further flexibility in the regularization or penalty term. This has the effect of allowing to trade the ability of the model to forget previous tasks (catastrophic forget) and its ability to learn new ones (intransigence).

\textbf{Uncertainty Estimation in Deep Learning.}

Despite its groundbreaking achievements, most deep learning models are not able to represent uncertainty. Although this might seem irrelevant in tasks such as distinguishing between cats and dogs, this is of key importance in more sensitive tasks such as those involving healthcare applications or those related with autonomous vehicles. Bayesian Neural Networks (BNNs)~\cite{neal1995bayesian}, on the other hand, are probabilistic models that produce uncertainty estimates by placing probability distributions over the weights, instead of just learning the most likely parameters given the training data. Unfortunately, modeling the full posterior distribution for the parameters of the model given the data is usually computationally intractable. Luckily, in the last decade, several variational techniques have been proposed to estimate the full posterior~\cite{graves2011practical,kingma2013auto,blundell2015weight}, so that making BNNs feasible. 

The novelty of our work lies in proposing a continual learning framework that, contrary to its predecessors, allows for further flexibility in the way it prioritizes the remembering of previous tasks or the learning of new ones, as well as in the application of such framework in the context of sensing-based analysis tasks.  

\section{Continual Learning Framework for Sensing-based Analysis Tasks}
\label{sec:methodology}

We present a framework to incrementally learn tasks from streams of sensory type data. Apart from mapping high dimensional data to an array of outputs, the learned representations also include uncertainty estimates that model the confidence on the predictions, which can then be used to accept or discard the predictions.

Below, we detail the algorithms behind our variational continual learning framework for sensing-based analysis tasks. Specifically, Section~\ref{sec:continual} describes the variational continual learning framework in~\cite{nguyen2018variational}~(\S~\ref{sec:VCL_model}), and how we extend it to allow for more flexibility in trading off the ability of the model to not forget previously learned tasks and to learn new ones~(\S~\ref{sec:beta_div}). Section~\ref{sec:ppd} then explains how we can use the learned model to perform inference or prediction on a new test input both for regression and classification. Finally, Section~\ref{sec:uncertainty} describes two different metrics to quantify the uncertainty of the prediction, namely entropy and mutual information.

\subsection{Continual learning by approximate Bayesian inference with $\beta$-divergence}
\label{sec:continual}

Our continual learning framework is an extension of the variational continual learning (VCL) framework for deep discriminative models proposed by Nguyen \emph{et al.} in~\cite{nguyen2018variational}. We make two variations to this framework. Firstly, due to the common storage limitations in edge computing environments, we consider the simple VCL model without coresets. That is, we do not keep samples of previous tasks that would presumably help prevent the model from forgetting older tasks. Secondly, we add a $\beta$ coefficient to the KL-divergence term in the loss function~(\S~\ref{sec:beta_div}). This coefficient is a hyperparameter in the model and, as we show in the evaluation, it helps to trade accuracy by \emph{forgetting}, \emph{intransigence} and vice versa.

\subsubsection{Variational Continual Learning (VCL) for deep discriminative models~\cite{nguyen2018variational}} 
\label{sec:VCL_model}
Consider a discriminative model that returns a probability distribution $p(y|\bm{\theta},\bm{x})$ over an output $y$ given an input $\bm{x}$ and parameters $\bm{\theta}$. The goal in continual learning is to learn the parameters of the model from a set of sequentially arriving datasets $\{\bm{x}_t^{(n)}, y_t^{(n)} \}_{n=1}^{N_t}$, where, in principle, each might contain a single datum, $N_t=1$. Following a Bayesian approach, a prior distribution $p(\bm{\theta})$ is placed over $\bm{\theta}$. The posterior distribution after seeing $T$ datasets is recovered by applying Bayes' rule:

\begin{equation}
	\begin{split}
p(\bm{\theta}|D_{1:T}) & \propto p(\bm{\theta}) \prod_{t=1}^{T} \prod_{n_t=1}^{N_t} p(y_t^{(n_t)}|\bm{\theta},\bm{x}_t^{(n_t)}) \\ & = p(\bm{\theta}) \prod_{t=1}^{T} p(D_t|\bm{\theta}) \propto p(\bm{\theta}|D_{1:T-1}) p(D_T|\bm{\theta}).
	\end{split}
\end{equation}

We have suppressed the input dependence on the right hand side to lighten the notation, and used the shorthand $D_t = \{y_t^{(n)}\}_{n=1}^{N_t}$. We note a recursion whereby the posterior, after seeing the $T$-th dataset, is produced by taking the posterior after seeing the $(T-1)$-th dataset, multiplying by the likelihood and renormalizing. In most cases, the posterior recursion $p(\bm{\theta}|D_t)$ at time $t$ is intractable and we should resort to approximations ($p(\bm{\theta}|D_t) \approx q_t(\bm{\theta})$). As in~\cite{nguyen2018variational}, we approximate the posterior through a KL divergence minimization over the set of allowed approximate posteriors $Q$.

\begin{equation}
	\begin{split}
q_{t}(\bm{\theta}) = \underset{q \in Q}{\arg\min} \text{KL} & \Big( \infdivx {q(\bm{\theta})} {\frac{1}{Z_t} q_{t-1}(\bm{\theta}) p(D_t|\bm{\theta})} \Big),  \\ & \text{ for } t=1,2,...,T.
	\end{split}
\end{equation}

\begin{sloppypar}
The zeroth approximate distribution is defined to be the prior, $q_0(\bm{\theta}) = p(\bm{\theta})$. $Z_t$ is the intractable normalizing constant of $q_{t-1}(\bm{\theta}) p(D_t|\bm{\theta})$ and it is not required to compute the optimum. 
\end{sloppypar}
	
We apply this framework to continually learn a deep fully-connected neural network classifier. In the case of multiple audio, activity or other sensing analysis tasks with different outputs, we consider a network that share parameters close to the inputs but with separate heads for each output (\emph{multi-head} network). 

\begin{sloppypar}
VCL requires to specify $q(\bm{\theta})$, where $\bm{\theta}$ is a $D$ dimensional vector formed by stacking the network's biases and weights. We use a Gaussian mean-field approximate posterior $q_t(\bm{\theta}) = \prod_{d=1}^{D} \mathcal{N} (\theta_{t,d}; \mu_{t,d}, \sigma_{t,d}^2)$. Before task $k$ is encountered the posterior distribution over the associated head parameters remains at the prior and so $q(\bm{\theta\_k^H}) = p(\bm{\theta\_k^H})$. Also, only tasks present in the current dataset $D_t$ have their posterior distributions over head parameters updated, although the shared parameters will be constantly updated. Finally, training the network is equivalent to maximizing the variational lower bound to the online marginal likelihood
\end{sloppypar}
	
\begin{dmath}
	\mathcal{L}_{VCL}^t (q_t(\bm\theta)) = \sum_{n=1}^{N_t} \mathbb{E}_{\bm{\theta} \sim q_t(\bm{\theta})} \big[ \text{log } p(y_t^{(n)}|\bm{\theta}, \bm{x}_t^{(n)}) \big]  - \text{KL} (\infdivx {q_t(\bm{\theta})} {q_{t-1}(\bm{\theta})})
	\label{eq:loss}
\end{dmath}

\noindent with respect to the variational parameters $\{ \mu_{t,d}, \sigma_{t,d} \}_{d=1}^D $. Whilst the KL-divergence $\text{KL} (\infdivx {q_t(\bm{\theta})} {q_{t-1}(\bm{\theta})})$ can be computed in closed-form, the expected log-likelihood requires further approximation. As in~\cite{nguyen2018variational}, we use Monte Carlo and the \emph{local reparametrization trick}~\cite{kingma2015variational} to compute the gradients. At the first time step, the prior distribution (and $q_0(\bm{\theta})$) is chosen to be a multivariate Gaussian.

\subsubsection{VCL with weighted $\beta$-divergence} 
\label{sec:beta_div}

Maximizing the loss in Eq.~\eqref{eq:loss} is equivalent to simultaneously maximizing the online marginal likelihood while minimizing the KL-divergence between the approximate posterior in $t$, $q_t(\bm{\theta})$, and the approximate posterior in the previous iteration $t-1$, $q_{t-1}(\bm{\theta})$. The latter has strong influence on avoiding \emph{catastrophic forgetting} since it prevents the parameters of the model $\bm{\theta}$ from deviating much from their value in the previous iteration. However, its influence is also notable in the inability of the model to learn new tasks (\emph{intransigence}), since the closer the parameters need to be to their previous values, the less freedom the model will have to learn new tasks. To give more flexibility to the model so that it can prioritize minimizing catastrophic forgetting or minimizing intransigence, we add a new hyperparameter $\beta \in [0,1]$ that controls the effect of the KL-divergence or \emph{penalty} term in the loss. By adding the $\beta$ term to Eq.~\eqref{eq:loss}, training the network is equivalent to maximizing the $\beta$-weighted variational lower bound to the online marginal likelihood

\begin{dmath}
	\mathcal{L}_{VCL}^t (q_t(\bm\theta)) = \sum_{n=1}^{N_t} \mathbb{E}_{\bm{\theta} \sim q_t(\bm{\theta})} \big[ \text{log } p(y_t^{(n)}|\bm{\theta}, \bm{x}_t^{(n)}) \big]  - \beta \text{KL} (\infdivx {q_t(\bm{\theta})} {q_{t-1}(\bm{\theta})}).
	\label{eq:loss_beta}
\end{dmath}

\subsection{Posterior probability distribution}
\label{sec:ppd}

In order to perform prediction on a new test input $\bm{x}_t^*$ for task $t$ we calculate the posterior probability distribution $p(y^*_t| \bm{x}^*_t, D_{1:T})$ by marginalizing out the posterior distribution: 
\begin{equation}
	\begin{split}
	p(y_t^* | \bm{x}_t^*, D_{1:T}) & =  \int_{-\infty}^{\infty} p(y_t^* | \bm{x}_t^*, \bm{\theta}) p(\bm{\theta} | D_{1:T}) d\bm{\theta} \\ & \approx \int_{-\infty}^{\infty} p(y_t^* | \bm{x}_t^*, \bm{\theta}) q_t(\bm{\theta}) d\bm{\theta} .
	\end{split}
	\label{eq:predPosterior}
\end{equation}

\noindent For classification, this can be approximated using Monte Carlo integration as follows~\cite{kendall2017uncertainties}:

\begin{multline}
	\bm{p}(y_t^*| \bm{x}_t^*, D_{1:T}) = [p(y_t^* = 1 | \bm{x}_t^*, D_{1:T}), \dotsc, \\ p(y_t^* = C | \bm{x}_t^*, D_{1:T})]^T  \approx \frac{1}{S} \sum_{s=1}^{S}  \text{Softmax}(\bm{f}^{\widehat{\bm{\theta}}_{t,s}}(\bm{x}_t^*))
	\label{eq:predPosteriorMC}
\end{multline}

\noindent with $S$ sampled masked model weights $\widehat{\bm{\theta}}_{t,s} \sim q_{t}(\bm{\theta})$, and $\bm{f}^{\widehat{\bm{\theta}}_{t,s}}(\bm{x}_t^*)$ the random output of the Bayesian Neural Network for the input $\bm{x}_t^*$. We then predict the class $c_t^*$ by 
\begin{equation} 
	c^* = \underset{c=1,\dotsc,C}{\arg\max} \ \bm{p}(y_t^*| \bm{x}_t^*, D_{1:T}).
\end{equation}

\subsection{Uncertainty quantification}
\label{sec:uncertainty}

Uncertainty computation is essential to the design of robust and reliable systems. In regression, the predictive uncertainty can be summarized by the sample variance of multiple stochastic forward passes. In a classification setting, however, we need to use alternative approaches to model uncertainty. Following the recommendations in~\cite{Gal2016UncertaintyID}, we use the predictive entropy as an indication of the \emph{uncertainty of the prediction}, and the mutual information between the prediction $y_t^*$ and the posterior over the model parameters $\bm{\theta}$ to capture the \emph{confidence of the model in its output}. 

The predictive entropy captures the average amount of information in the predictive distribution. It is computed as follows

\begin{equation}
	\begin{split}
	\mathbb{H}[y_t^*| \bm{x}_t^*, D_{1:T}] = -\sum_c & p(y_t^* = c | \bm{x}_t^*, D_{1:T}) \\ & \log{p(y_t^* = c | \bm{x}_t^*, D_{1:T})}
	\end{split}
\end{equation}

The mutual information between prediction $y_t^*$ and the posterior over the model parameters $\bm{\theta}$ captures the mutual dependence between the two variables. It is computed as follows

\begin{equation}
		\mathbb{I}[y_t^*, \bm{\theta} | \bm{x}_t^*, D_{1:T}] = \mathbb{H}[y_t^*| \bm{x}_t^*, D_{1:T}] - \mathbb{E}_{p(\bm{\theta} | D_{1:T})} [ \mathbb{H}[y_t^* | \bm{x}_t^*, \bm{\theta}] ] 
\end{equation}
	
\noindent where the second term can be approximated by
 
\begin{dmath}
	\mathbb{E}_{p(\bm{\theta} | D_{1:T})} [ \mathbb{H}[y_t^* | \bm{x}_t^*, \bm{\theta}] ] \approx - \frac{1}{S} \sum_{c,s} p(y_t^* = c | \bm{x}_t^*, \widehat{\bm{\theta}}_{t,s}) \log{p(y_t^* = c | \bm{x}_t^*, \widehat{\bm{\theta}}_{t,s})}	 	
\end{dmath}

\noindent with 

\noindent $[p(y_t^* = 1 | \bm{x}_t^*, \widehat{\bm{\theta}}_{t,s}), \dotsc, p(y_t^* = C | \bm{x}_t^*, \widehat{\bm{\theta}}_{t,s})]^T = \text{Softmax}(\bm{f}^{\widehat{\bm{\theta}}_{t,s}}(\bm{x}_t^*))$.
\section{Evaluation: tasks, metrics and experimental setup}
\label{sec:expSetup}

We now detail how we applied our continual learning framework to learn some of the most common tasks that run on off-the-shelf mobile and edge devices. Specifically, we continually learn a model to classify multiple audio tasks and another one to classify activity-related ones. We describe the tasks and datasets considered (\S\ref{sec:audio} and \S\ref{sec:activity}), the performance metrics to evaluate the models (\S\ref{sec:metrics}), and the experimental setup (\S\ref{sec:expSetupSub}).

\subsection{Audio analysis tasks}
\label{sec:audio}

We consider four popular audio tasks namely speaker identification, ambient scene detection, stress detection and emotion recognition, in our experiments. Four different large-scale or widely adopted datasets were used in our evaluation. Below, we detail these tasks and datasets.

	\textbf{Speakers Identification (SI)}. The aim of this task is to identify the current speaker based on speech audio data. We consider 10-minute speech samples recorded by a total of $23$ speakers working in the Department of Computer Science and Technology of the University of Cambridge in 2013~\cite{georgiev2014dsp}. We used audio samples of 5 seconds of speech from a speaker as this duration has been used for mobile sensing tasks in social psychology experiments~\cite{rachuri2010emotionsense,georgiev2014dsp}. 
	
	\textbf{Ambient Scene Detection (ASD)}. The dataset consists of 40 minutes of various sounds equally split into the 4 categories music, traffic, water and other as in~\cite{georgiev2014dsp}. The music audio clips are a subset of the GTZAN genre collection~\cite{tzanetakis2002musical}; the traffic samples were downloaded from an online provider of free sound effects~\cite{FSE}; the water samples were obtained from the British Library of Sounds~\cite{BLS}; the rest of the sounds were crawled from a subset of the SFX dataset~\cite{chechik2008large}. Each sound scene is 1.28 seconds long, a commonly adopted window in mobile audio sensing~\cite{georgiev2017low}. 
	
	\textbf{Stress Detection (SD)}. The goal of this task is to detect stressed speech. We use a 1-hour recording of stressed and neutral speech, which is a subset of the dataset collected in \emph{StressSense}~\cite{lu2012stresssense}. We consider inference windows of length $1.28$ as in~\cite{lu2012stresssense}. 
	
	\textbf{Emotion Recognition (ER)}. This task aims to recognize 5 emotional categories from voice, namely \emph{neutral}, \emph{happy}, \emph{sad}, \emph{angry} and \emph{frightened}. We used the 2.5 hours of emotional speech delivered by professional actors in the Emotional Prosody Speech and Transcripts corpus~\cite{liberman2002emotional}. As in Rachuri \emph{et al.}~\cite{rachuri2010emotionsense}, we grouped the 14 narrow emotions into the above 5 categories. In our case, we consider inference windows of length 5 seconds. 

\subsubsection{Input audio features}
\label{sec:audioF}
We unify the feature extraction process across the different audio analysis tasks in order to have a shared feature representation. We use the log filter banks, an early step in the computation pipelines of the Perceptual Linear Predictive (PLP) coefficients and Mel Frequency Cepstral Coefficients (MFCC). We go a step further by using summaries of filter bank coefficients since, as Georgiev \emph{et al.}~\cite{georgiev2017low} demonstrated, they have the benefit of requiring significantly fewer processing resources, which would be highly beneficial in edge and mobile computing setups. 

Following~\cite{georgiev2017low}, we extract $24$ filter banks from each audio frame over a time window of 30 ms (with 10 ms stride), and summarize the distribution of the values for each coefficient across successive frames within a large context window with statistical transformations (min, max, std, mean, median, 25-percentile and 75-percentile), resulting into 168 different features per input sample. We standarize the features across individual datasets to have zero mean and unit variance before feeding them into the classifier.

\subsection{Activity analysis tasks}
\label{sec:activity}
We now consider two different activity analysis tasks, namely human activity recognition (HAR) and person identification (PI). The aim of HAR is to identify activities that users perform using accelerometer data captured with a smartphone or wearable device, whereas PI aims to uniquely identify the person using the device.

We use the \emph{WISDM} Human Activity Recognition dataset~\cite{kwapisz2011activity}, which consists of accelerometer data from $36$ subjects performing 6 different activities (\emph{walking}, \emph{jogging}, \emph{walking upstairs}, \emph{walking downstairs}, \emph{sitting} and \emph{standing}). These subjects carried an Android phone in their front pants leg pocket while performed each one of these activities for specific periods of time. Various time domain variables were extracted from the signal, and we consider the statistical measures obtained for every 10 seconds of accelerometer samples (sampled at 20 Hz) in~\cite{kwapisz2011activity} as the $d = 43$ dimensional features in our models. We standarize the features to have zero mean and unit variance before feeding them into the classifier. 
  Our final sample contains $5,418$ accelerometer traces, with on average $150.50$ traces per user and standard deviation of $44.73$. We randomly select $18$ users for the HAR task and the remaining $18$ for the PI task.

\subsection{Continual learning metrics}
\label{sec:metrics}
As in Chaudhry \emph{et al.}~\cite{chaudhry2018riemannian}, we consider the average accuracy of the network, \emph{how much} the network forgets previous tasks and its intransigence or inability to learn new tasks after it was trained incrementally from task $1$ to $k$ as indicators of its performance. 

	\textbf{Average accuracy (A)}. Let $a_{k,j} \in [0,1]$ be the accuracy of the $j$-th task ($j \leq k$) after training the network incrementally from task $1$ to $k$. Then, the average accuracy at task $k$ is $A_k = \frac{1}{k} \sum_{j=1}^k a_{k,j}$. The higher the $A_k$, the better the classifier.
	 
	\textbf{Forget (F)}. We quantify the forget for the $j$-tasks after the model has been incrementally trained up to task $k > j$ as
	\begin{equation}
		f_j^k = \max_{l \in \{ 1,...,k-1 \}} a_{l,j} - a_{k,j}, \quad \forall j<k . 
	\end{equation}
	Then, the average forget at $k$-th task is written as $F_k = \frac{1}{k} \sum_{j=1}^{k-1} f_j^k$. Likewise, as we are interested in minimizing the forget for each task after each (re-)training, we define $F = \frac{1}{K} \sum_{k=1}^K F_k$. The lower the $F$, the better the classifier.
	
	\textbf{Intransigence (I)}. In order to quantify the inability of the model to learn new tasks, Chaudhry \emph{et al.}~\cite{chaudhry2018riemannian} propose to compare the current model with the multitask model that has access to all data at all times. As we are in an incremental learning setup we feel this is a very strong assumption and that we should instead quantify the inability of the model to learn by only having access to the tasks the model has seen so far. So, we use as the reference model for each task, the model trained with the dataset for each task in isolation. Therefore, for each task $k$, we train a reference model with dataset $D_k$ and measure its accuracy, denoted as $a_k^*$. We then define the intransigence for the $k$-th task as $I_k = a_k^* - a_{k,k}$ where $a_{k,k}$ denotes the accuracy on the $k$-th task when incrementally trained up to task $k$. The lower the $I_k$, the better the classifier.

\subsection{Experimental setup}
\label{sec:expSetupSub}

Although the trend in computer vision and NLP is to use progressively larger networks, the size of the models for audio sensing in edge computing environments is usually constrained by runtime and memory. We use model sizes comparable to other models employed in embedded settings: 3 hidden layers with 128 nodes each for keyword spotting~\cite{chen2014small}, 4 hidden layers with 256 nodes each for speaker verification~\cite{variani2014deep}, and 3 hidden layers with 512 nodes each for speaker identification, emotion recognition, stress detection and ambient scene analysis in a multitask learning setup~\cite{georgiev2017low}. We consider a network with 3 hidden layers and 512 nodes. 

We train our deep fully-connected neural network classifier for continually learning audio (activity) analysis tasks. We divide each dataset into training, validation and test sets with an $80\%-10\%-10\%$ split. At each (re-)training time $k$, we consider the training data of only one task, but the validation (test) data of all the tasks seeing since $t=1$ until $k$. We report average classification accuracy, average forget and average intransigence in the test set as described in~\S\ref{sec:metrics} as model performance metrics. We limit the total (re-)training time to 120 epochs across experiments, and sample the weights distribution $10$ times for the forwards pass during training and $100$ during inference/test time. The size of the input layer is $168$ in the audio model and $43$ in the activity one, corresponding to the dimensional features of each audio (activity) sample.
\section{Results}
\label{sec:evaluation}

We now detail our results. This evaluation focuses on three aspects. Firstly, we train our hyperparameters to maximize the average accuracy of the different audio and activity tasks, but at the same time minimizing the \emph{forget} and \emph{intransigence} of the trained models. In order to assess the robustness of this methodology, we perform this hyperparameter search using different orders for the audio (activity) tasks.
Secondly, we study the differences between two measures of uncertainty, namely \emph{entropy} and \emph{mutual information}, for those samples in the test set correctly and mistakenly classified. Thirdly, we analyze the feasibility of deploying our trained continual learning model in a mobile device. 

\subsection{Hyperparameter selection}
\label{sec:hyp}

In order to test the robustness of our hyperparameter search, we performed several experiments considering different orders of arriving tasks. For each order, we incrementally train a model using different learning rates ($0.0001$, $0.0005$, $0.001$, $0.005$, $0.01$) and $\beta$ parameters ($0.001$, $0.01$, $0.05$, $0.1$, $0.5$, $1.0$). Table~\ref{tab:hyp_audio} (Table~\ref{tab:hyp_har}) shows, for each of the 4 (2) different orders of audio (activity) tasks, the minimum, thus optimal, value of our combined performance metric, $(1-A_k) + F_k + I_k$, on the validation set after training the network incrementally from task $1$ to $k$, for all $k$. It also shows the average accuracy $A_k$, forget $F_k$ and intransigence $I_k$, as well as learning rate and parameter $\beta$ that minimize $(1-A_k) + F_k + I_k$. Note that sometimes the average intransigence is negative, i.e., the intransigence for some of the tasks is negative. This happens because these tasks benefit from the learning of other similar tasks.

\begin{table}
\small
\setlength{\tabcolsep}{2pt}
\begin{center}
\begin{tabular}{|l|l|c|c|c|c|c|c|c|c|}
\hline
\emph{Tasks order} & $k$ & $A_k$ & $F_k$ & $I_k$ & \multicolumn{1}{p{1.35cm}|}{\centering $(1-A_k) +$ \\ $F_k + I_k$} &  $l_{rate}$ & $\beta$ \\
\hline
\multirow{4}{*}{SI-ASD-SD-ER} & $1$ & $.9721$ & $.0$ & $.0$ & $.0279$ & $.01$ & $.01$\\
& $2$  & $.9284$ & $.008$ & $.0$ & $.0796$ & $.0005$ & $.1$ \\
& $3$  & $.8411$ & $.0467$ & $.0306$ & $.2362$ & $.001$  & $.1$ \\
& $4$  & $.78685$ & $.0866$ & $.0$ & $.2997$ & $.001$ & $.1$\\

\hline
\multirow{4}{*}{ER-SD-ASD-SI} & $1$ & $.8508$ & $.0$    & $.0$    & $.1492$ & $.005$ & $.001$ \\
							  & $2$ & $.7452$ & $.0282$ & $.0389$ & $.3219$ & $.001$ & $.5$ \\
							  & $3$ & $.7932$ & $.0499$ & $.0114$ & $.2681$ & $.001$ & $.1$ \\
							  & $4$ & $.7846$ & $.1056$ & $-.0080$ & $.3129$ & $.001$ & $.05$ \\

\hline
\multirow{4}{*}{SD-SI-ASD-ER} & $1$ & $.7917$ & $.0$ & $.0$ & $.2083$ & $.001$ & $.001$ \\
							  & $2$ & $.8572$ & $.0278$ & $-.0120$ & $.1585$ & $.001$ & $.05$ \\
							  & $3$ & $.8496$ & $.0456$ & $.0$ & $.1960$ & $.0005$ & $.1$ \\
							  & $4$ & $.8006$ & $.0894$ & $-.0080$ & $.2808$ & $.001$ & $.05$ \\

\hline
\multirow{4}{*}{ASD-SI-ER-SD} & $1$ &  $.9429$ & $.0$ & $.0$ & $.0571$ & $.01$ & $.1$ \\
							  & $2$ & $.9075$ & $.0571$ & $-.0080$ & $.1415$ & $.001$ & $.1$  \\
							  & $3$ & $.8168$ & $.0782$ & $.0201$ & $.2816$ & $.0005$ & $.1$ \\
							  & $4$ & $.7341$ & $.0297$ & $.0111$ & $.3066$ & $.0001$ & $1.0$  \\

\hline
\end{tabular}
\caption{Performance metrics of the best audio model after it was trained up to $k$ tasks evaluated on the validation set. Results are provided for different order of tasks.}
\label{tab:hyp_audio}
\end{center}
\end{table}

\begin{table}
\small
\setlength{\tabcolsep}{2pt}
\begin{center}
\begin{tabular}{|l|l|c|c|c|c|c|c|c|c|}
\hline
\emph{Tasks order} & $k$ & $A_k$ & $F_k$ & $I_k$ & $(1-A_k) + F_k + I_k$ &  $l_{rate}$ & $\beta$ \\
\hline
\multirow{2}{*}{HAR-PI} & $1$ & $.824$ & $.0$ & $.0$ & $.176$ & $.001$ & $.5$\\
& $2$  & $.732$ & $.149$ & $.027$ & $.445$ & $.001$ & $.1$ \\

\hline
\multirow{2}{*}{PI-HAR} & $1$ & $.871$ & $.0$    & $.0$    & $.129$ & $.01$ & $.05$ \\
						& $2$ & $.801$ & $.007$ & $.047$ & $.253$ & $.001$ & $1.0$ \\

\hline
\end{tabular}
\caption{Performance metrics of the best activity model after it was trained up to $k$ tasks evaluated on the validation set. Results are provided for different order of tasks.}
\label{tab:hyp_har}
\end{center}
\end{table}

We make the following observations in Tables~\ref{tab:hyp_audio} and~\ref{tab:hyp_har}. Firstly, as the accuracy of the different tasks when trained in isolation ($k=1$ in the Tables) varies a lot between tasks, the accuracy after training $k$ tasks is different in each one of the orders. Only after training up to $k=4$ ($k=2$), the average accuracy obtained gets closer. Secondly, the minimum value of $(1-A_k) + F_k + I_k$ gets higher as more tasks are trained. This happens mainly because the error $1-A_k$ increases when training more tasks. More importantly, the values of the optimal hyperparameters ($l_{rate}$ and $\beta$) are quite stable and similar when more than one task is learned. For instance, the best learning rate fluctuates between the consecutive values of $0.0005$ and $0.001$ when 2 or more audio tasks are learned, and is stable and equal to $0.001$ in the activity model. The best $\beta$ in the audio model fluctuates between $0.05$ and $0.5$, but mostly stays in $0.1$, whereas the optimal value of $\beta$ fluctuate between $0.05$ and $1.0$ in the activity case. This means that \emph{data from any 2 tasks is often enough to learn the optimal hyperparameters to use for the continual learning model on audio analysis tasks}.

\begin{figure}[!ht]
  \centering
  \includegraphics[scale=0.55]{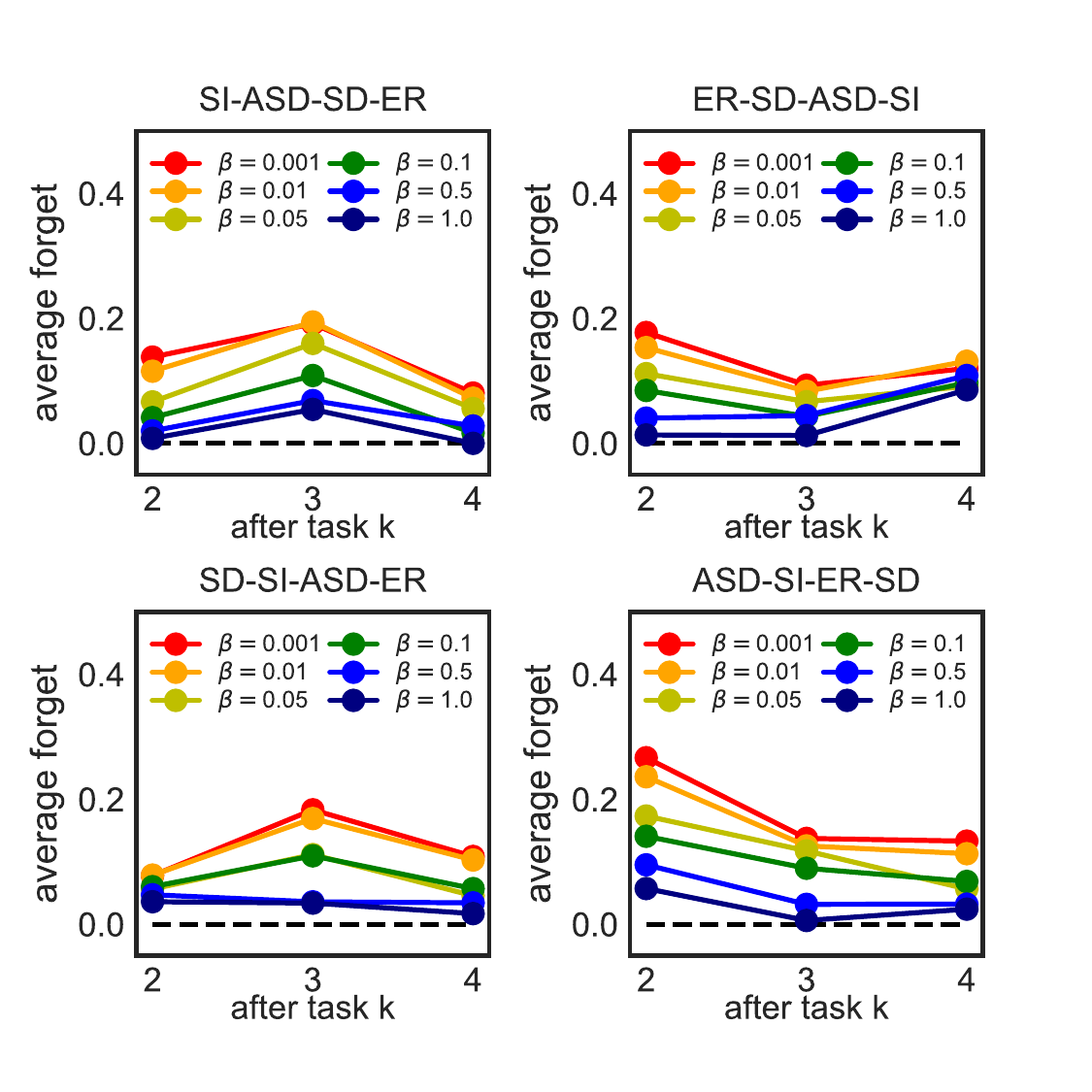}
  \caption{Average forget after training the audio model incrementally up to task $k$ for 4 different orders of tasks and different $\beta$s.}
  \label{fig:beta_f}
\end{figure}

\begin{figure}[!ht]
  \centering
  \includegraphics[scale=0.55]{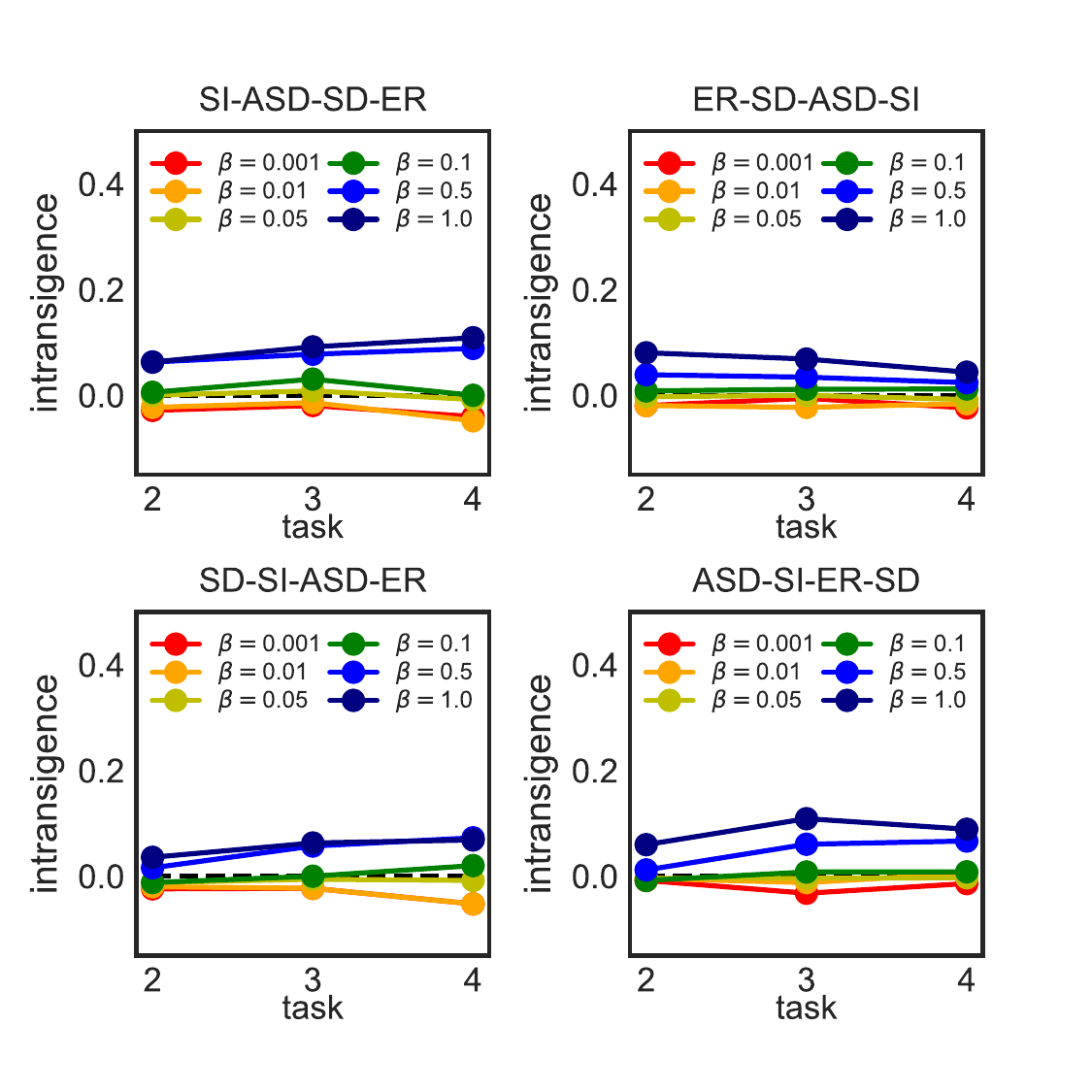}
  \caption{Intransigence of each task after training the audio model incrementally up to task $k$ for 4 different orders of tasks and $\beta$s.}
  \label{fig:beta_i}
\end{figure}

\textbf{Effect of $\beta$ on forget and intransigence}. 
We now fix the learning rate to $0.001$ (both for the audio and activity models) and plot (show), in Figures~\ref{fig:beta_f} and~\ref{fig:beta_i} (Table~\ref{tab:forget_int_activity}), the average forget and intransigence when using different $\beta$ parameters for the different orders of tasks evaluated on the validation set. In both cases, we observe that, on one hand, for higher $\beta$s, and therefore when more importance the prior has on the posterior, the less the model forgets previous tasks. On the other hand, with lower values of $\beta$, the ability of the model to learn new tasks (intransigence) increases. Therefore, \emph{$\beta$ could be tuned in order to build a model that either minimizes the forget of old tasks or minimizes the intransigence of new tasks}.

\begin{figure}[!ht]
  \centering
  \includegraphics[scale=0.55]{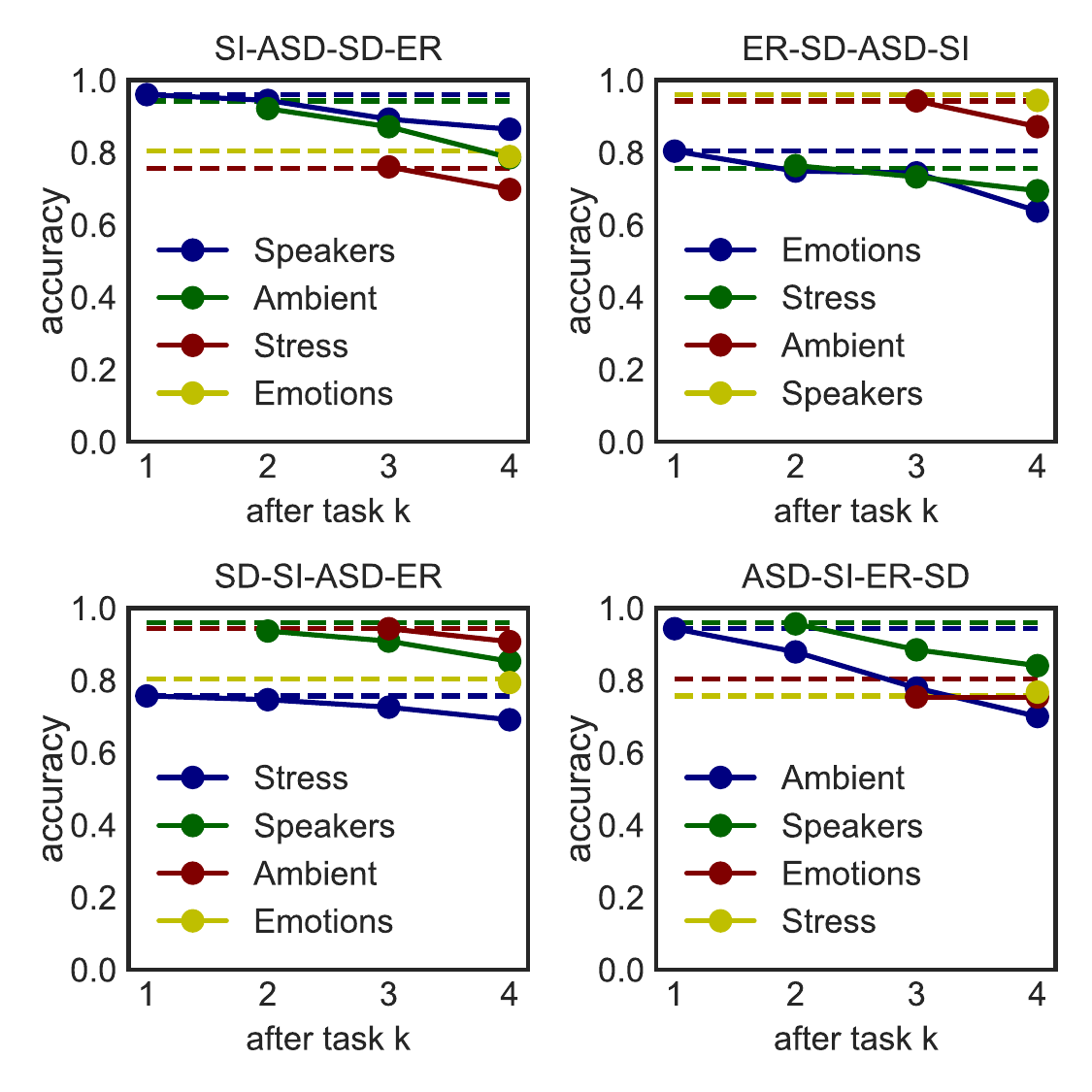}
  \caption{Accuracy for each task after training the audio model incrementally up to task $k$ for 4 different orders of tasks. Dashed lines indicate the accuracy when trained the model in isolation.}
  \label{fig:acc_per_task}
\end{figure}

\begin{figure}[!ht]
  \centering
  \includegraphics[scale=0.55]{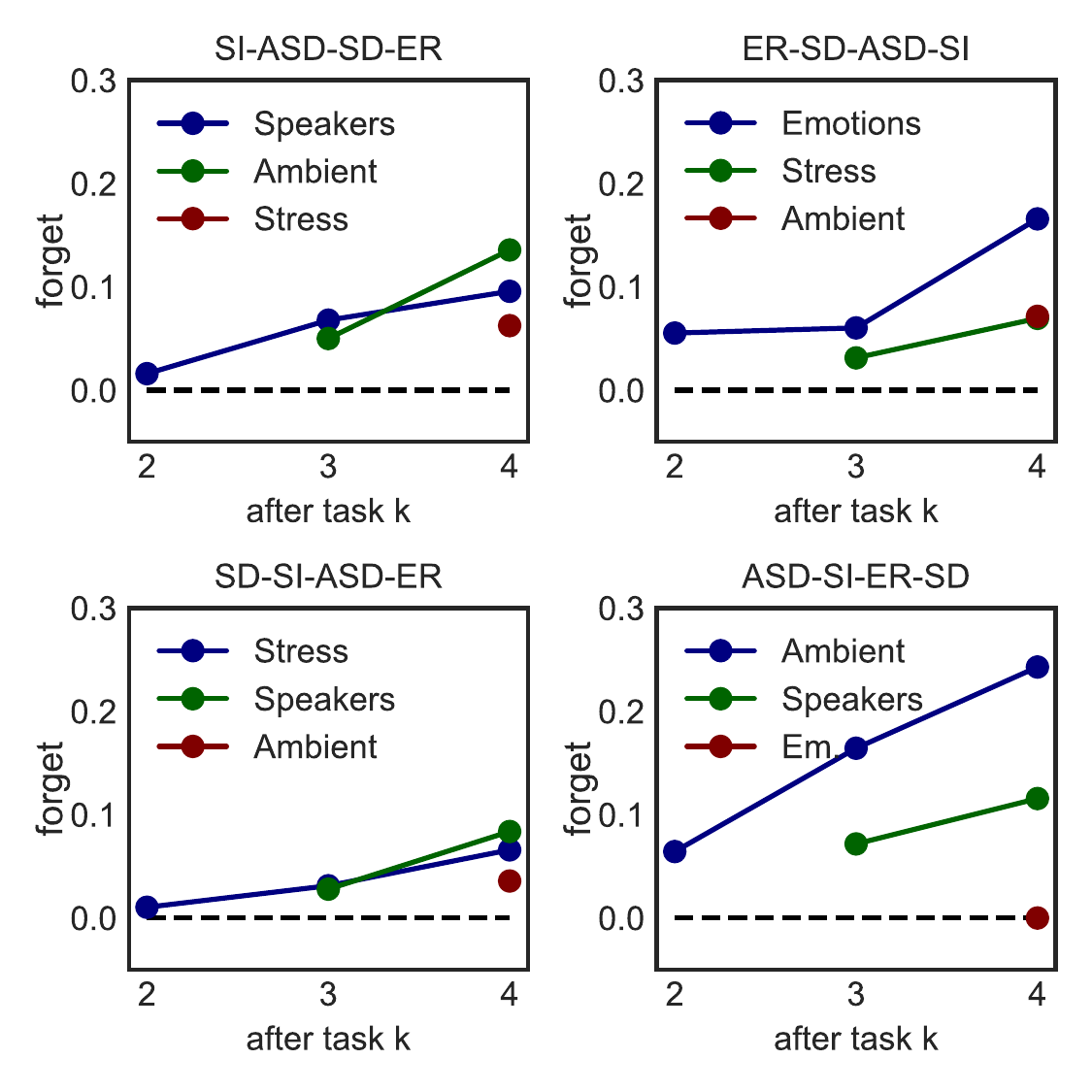}
  \caption{Forget for each task after training the audio model incrementally up to task $k$ for 4 different orders of tasks. The dashed line indicates the ideal forget value of $0$.}
  \label{fig:forg_per_task}
\end{figure}

\begin{figure}[!ht]
	        \centering
	        \subfloat[]{\label{fig:int_per_task}\includegraphics[scale=0.4]{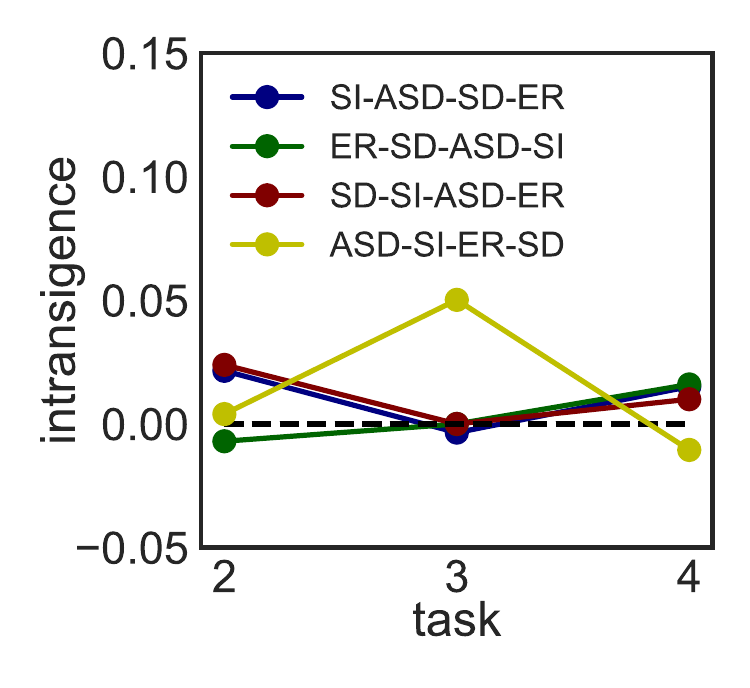}} 
	        \hfill 			\subfloat[]{\label{fig:acc_per_task_activity}\includegraphics[scale=0.5]{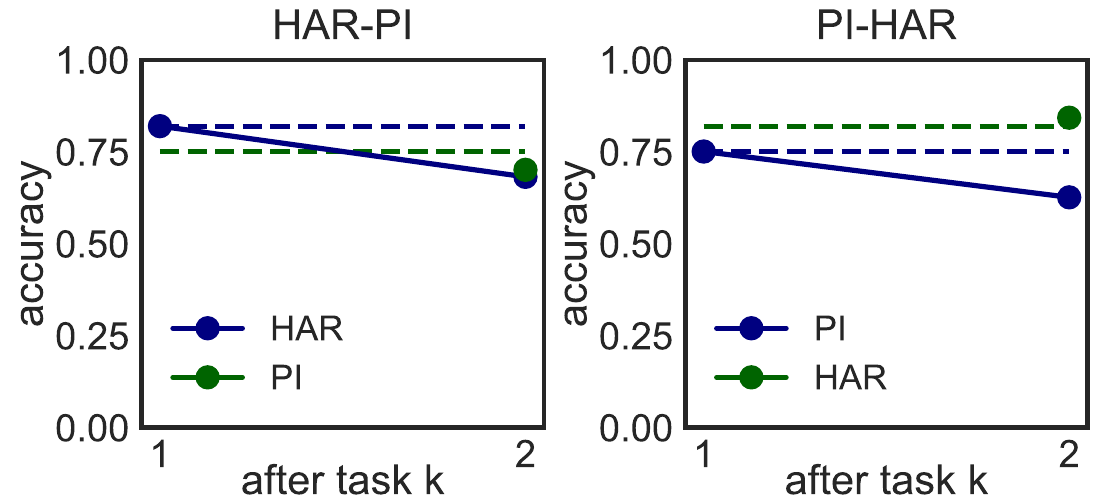}}
	        \caption{(a) Intransigence for each task after training the audio model incrementally up to task $k$ for 4 different orders of tasks. The dashed line indicates the ideal intransigence value of $0$. (b) Accuracy for each task after training the activity model incrementally up to task $k$. Dashed lines indicate the accuracy when trained the model in isolation for each task.}
	          \label{fig:int_acc_both_tasks}
	\end{figure}

\textbf{Results on the test set}. We select the models trained with $l_{rate} = 0.001$ and $\beta = 0.1$ as the ones to use in the rest of the experiments. Figures~\ref{fig:acc_per_task}, \ref{fig:forg_per_task} and \ref{fig:int_per_task} show the accuracy, forget and intransigence for each one of the audio tasks and the different orders evaluated on the test set. Similarly, Figure~\ref{fig:acc_per_task_activity} shows how the accuracy varies for each activity task when learning new tasks. We also obtained that the forget of the second task in the order HAR-PI is $0.138$ and $0.125$ when the order is PI-HAR, whereas the intransigence of the model is very low in both cases: $0.050$ when HAR is trained first and $-0.023$ when PI is the first task learned.

We observe that accuracy, forget and intransigence values are similar to the ones in the validation set. More importantly, the order in which tasks arrive have a strong influence in the accuracy, forget and intransigence for each task after the models have been trained up to task $k$. As for accuracy, we also find that, the optimal hyperparameters selected according to our metric are those that give more importance to minimize intransigence than forget. Note, though, that a different optimization metric could be used if we were more interested in, for example, minimizing the forget of the previous tasks than on minimizing the intransigence of the model.

\textbf{Comparison with baselines}. We consider \emph{single task models} and the \emph{continual learning model with $\beta = 1$} as baselines. Accuracy values obtained on the test set for single task models correspond to the accuracy \emph{after task k=1} in Fig.~\ref{fig:acc_per_task} and Fig.~\ref{fig:acc_per_task_activity}. As expected, the accuracy of the continual learning model is usually lower than that of the single task one as it needs to \emph{accommodate} the learning of new tasks. The exception is when the task considered is the last task to be learned in the continual learning setup, as this can also benefit from the knowledge gained by previously learned tasks.
 Moreover, although the metric we use to search for the optimal hyperparameters ($(1-A_k) + F_k + I_k$) seems to prioritize the accuracy of new tasks over the remembering of old ones, it can also be modified in order to prioritize the accuracy of selected tasks over others by tuning the $\beta$ parameter accordingly. In line with the latter, we observe that the model with $\beta=1$ (corresponding to $\beta=1$ in Fig.~\ref{fig:beta_f} and Fig.~\ref{fig:beta_i}) prioritizes the remembering of old tasks (forget) over the ability to learn new ones (intransigence), and thus it is more accurate than a model with $\beta \neq 1$ for previous tasks, but less accurate for newly learned ones.

\begin{table}
\scriptsize
\setlength{\tabcolsep}{1.5pt}
\begin{center}
\begin{tabular}{|l|c|c|c|c|c|c|c|c|c|c|c|c|}
\hline
& \multicolumn{6}{c|}{$F_k$} & \multicolumn{6}{c|}{$I_k$} \\
\hline
\emph{$\beta$} & $.001$ & $.01$ & $.05$ & $.1$ & $.5$ & $1.0$ & $.001$ & $.01$ & $.05$ & $.1$ & $.5$ & $1.0$ \\
\hline
{HAR-PI} & $.223$ & $.211$ & $.172$ & $.149$ & $.176$ & $.134$ &
$-.018$ & $-.007$ & $.021$ & $.036$ & $.150$ & $.214$ \\
\hline
{PI-HAR}  & $.218$ & $.186$ & $.107$ & $.093$ & $.036$ & $.007$ & 
$-.007$ & $.004$ & $.011$ & $.023$ & $.019$ & $.042$ \\
\hline
\end{tabular}
\caption{Forget and intransigence of the second activity task after training the model incrementally up to task $k$ for 2 different orders of tasks and different values of $\beta$.}
\label{tab:forget_int_activity}
\end{center}
\end{table}

\subsection{Classification uncertainty}

We now compute the entropy $\mathbb{H}[y_t^*| \bm{x}_t^*, D_{1:T}]$ and mutual information $\mathbb{I}[y_t^*, \bm{\theta} | \bm{x}_t^*, D_{1:T}]$ for each data point in the test set of each audio task $t \in {1..K}$ (activity task $t \in {1..2}$) after training the audio model (activity model) incrementally up to task $t \in {1..4}$ ($t \in {1..2}$) for the tasks order SI-ASD-SD-ER (HAR-PI). 

\begin{figure}[!ht]
  \centering
  \includegraphics[scale=0.5]{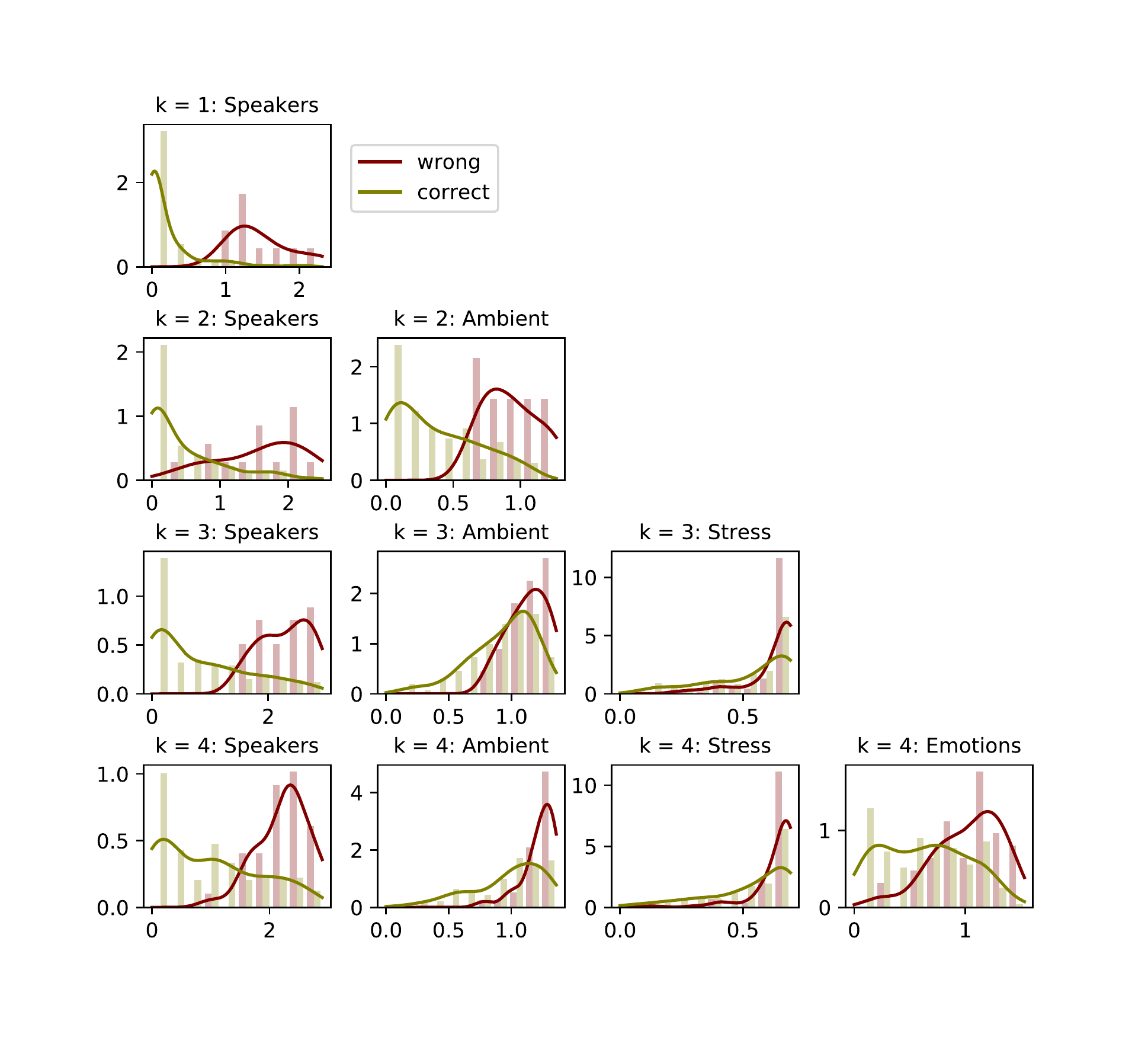}
  \caption{Probability distribution and probability density estimate of the entropy for each task after training the audio model incrementally up to task $k$ for the task order SI-ASD-SD-ER.}
  \label{fig:entropy}
\end{figure}

\begin{figure}[!ht]
  \centering
  \includegraphics[scale=0.5]{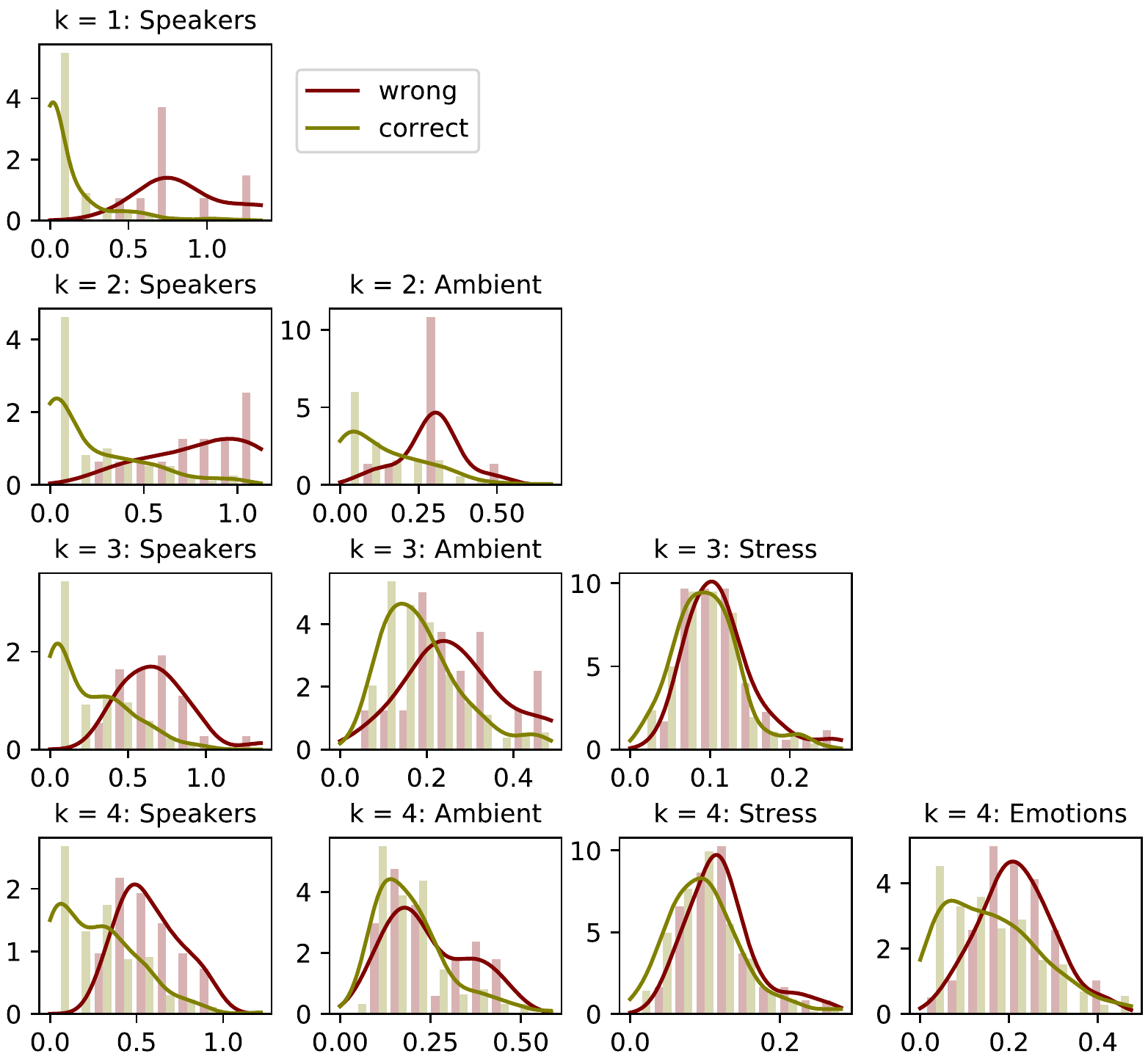}
  \caption{Probability distribution and probability density estimate of the mutual information for each task after training the audio model incrementally up to task $k$ for the task order SI-ASD-SD-ER.}
  \label{fig:MI}
\end{figure}

\begin{figure}[!ht]
	        \centering
	        \subfloat[]{\label{fig:entropy_act:a}\includegraphics[scale=0.53, angle=0]{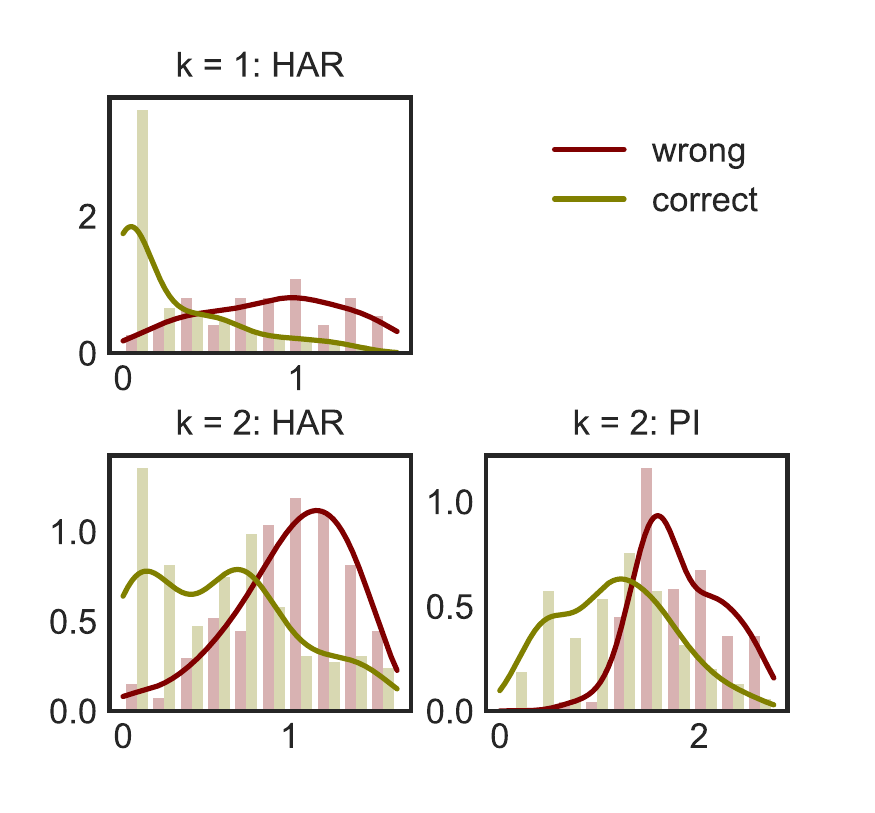}} 
	        \hfill
	        \subfloat[]{\label{fig:entropy_act:b}\includegraphics[scale=0.53, angle=0]{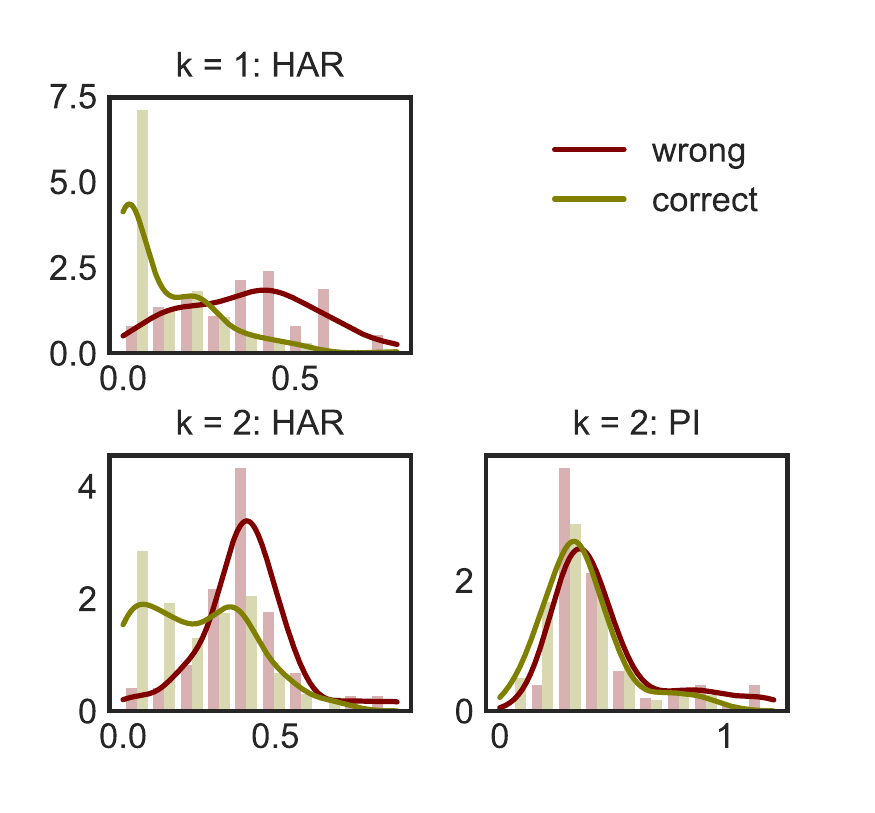}} 
	        \caption{Probability distributions and probability density estimates of the (a) entropy and (b) mutual information for each activity analysis task after training the model incrementally up to task $k$ for the task order HAR-PI.}
	          \label{fig:entropy_act}
	\end{figure}

 Figures~\ref{fig:entropy} and \ref{fig:MI} (Figure~\ref{fig:entropy_act}) show the distributions of the entropy and mutual information values obtained with the audio (activity) model. We observe that, in general, the data points with higher entropy and higher mutual information are more prone to be classified wrongly. This is especially true for the entropy. Also, as $k$ increases, i.e., the model is trained to learn more tasks, the differences between the uncertainty of those samples wrongly classified and those correctly classified reduces. More importantly, the uncertainty increases when adding more tasks to the model. 
		
\begin{table}
	\small
	\setlength{\tabcolsep}{2pt}
	\begin{center}
		
     \begin{tabular}{|l|l|c|c|c|c|c|}
     \hline
     & & \multicolumn{2}{c|}{\emph{entropy}} & \multicolumn{2}{c|}{\emph{mutual info.}} \\
     \hline
     \emph{Tasks order} & $k$ & $t_1$ & $t_2$ & $t_1$ &  $t_2$ \\
     \hline
     \multirow{2}{*}{HAR-PI}  & $1$ & $.000$ & & $.000$ & \\
     							   & $2$ & $.000$ & $.000$ & $.000$ & $.002$ \\

     \hline
     \multirow{2}{*}{PI-HAR}  & $1$ & $.000$ & & $.000$ & \\
     							   & $2$ & $.000$ & $.000$ & $.000$ & $.000$ \\
     \hline
     \end{tabular}
        \caption{$p$-values resulting from the Kruskal-Wallis test on the activity model. Results are provided for different order of tasks.}
		\label{tab:kruskal_activity}
		\end{center}
      \end{table}

\textbf{Significance of the findings}. We now test, for each possible order of tasks, trained model and task, whether the entropy (mutual information) of the samples correctly classified is different to the entropy (mutual information) of the samples wrongly classified. To do so, we first run the Kolmogorov-Smirnov Test to check the normality of our distributions, obtaining $p<0.05$ in all the cases. Thus we do not believe that the entropy (mutual information) of the samples correctly classified nor of the samples wrongly classified follows a normal distribution. We then used the non parametric Kruskal-Wallis Test to test, for each model and task, the null hypothesis that the entropy (mutual information) median of the wrongly classified samples is equal to the median of those correctly classified. Tables~\ref{tab:kruskal} and~\ref{tab:kruskal_activity} show the resulting $p$-values for the audio and activity experiments, respectively. 

Considering the entropy of the prediction, the $p$-values are lower than $0.05$ for all the orders, tasks and trained models, which means that we can reject the null hypothesis and therefore this proves that the median entropy of the samples wrongly classified is indeed different than the one of those correctly classified. Moreover, as we observe in Figure~\ref{fig:entropy} and Figure~\ref{fig:entropy_act:a}, wrongly classified samples tend to have higher entropy than those correctly classified. Similarly, for the mutual information, we can reject the null hypothesis for most of the orders, tasks and trained models. The exception is for the stress detection (audio) task in some of the orders tested, where $p$ is higher than $0.05$ and we cannot conclude that the mutual information median of the wrongly classified samples is different than the median of those correctly classified. As with entropy, Figure~\ref{fig:MI} and Figure~\ref{fig:entropy_act:b} show that 
 the mutual information tends to be higher in the samples wrongly classified than in those correctly classified. Therefore, \emph{entropy and mutual information might be used to decide which predictions to trust and which ones to discard}.

\begin{table}
\small
\setlength{\tabcolsep}{2pt}
\begin{center}
\begin{tabular}{|l|l|c|c|c|c|c|c|c|c|c|}
\hline
& & \multicolumn{4}{c|}{\emph{entropy}} & \multicolumn{4}{c|}{\emph{mutual information}} \\
\hline
\emph{Tasks order} & $k$ & $t_1$ & $t_2$ & $t_3$ & $t_4$ & $t_1$ &  $t_2$ & $t_3$ & $t_4$ \\
\hline
\multirow{4}{*}{SI-ASD-SD-ER} & $1$ & $.000$ & & & & $.000$ & & & \\
  & $2$& $.000$ & $.000$ & & & $.000$ & $.000$  & & \\
  & $3$& $.000$ & $.002$ & $.000$ & & $.000$ & $.002$ & $.009$ & \\
  & $4$& $.000$ & $.000$ & $.000$ & $.000$ & $.000$ & $.045$ & $.000$ & $.000$ \\

\hline
\multirow{4}{*}{ER-SD-ASD-SI}  & $1$ & $.000$ & & & & $.000$ & & & \\
							   & $2$ & $.000$ & $.000$ & & & $\mathbf{.054}$ & $.000$ & & \\
							   & $3$ & $.000$ & $.000$ & $.000$ & & $.002$ & $.000$ & $.001$ & \\
							   & $4$ & $.000$ & $.000$ & $.000$ & $.000$ & $.006$ & $.003$ & $.000$ & $.000$ \\

\hline
\multirow{4}{*}{SD-SI-ASD-ER}  & $1$ & $.000$ & & & & $.000$ & & & \\
							   & $2$ & $.000$ & $.000$ & & & $.000$ & $.000$ & & \\
							   & $3$ & $.000$ & $.000$ & $.000$ & & $.000$ & $.000$ &  $.001$ & \\
							   & $4$ & $.000$ & $.000$ & $.000$ & $.000$ & $.000$ & $.000$ & $.000$ & $.000$ \\

\hline
\multirow{4}{*}{ASD-SI-ER-SD}  & $1$ & $.000$ & & & & $.000$ & & & \\
							   & $2$ & $.000$ & $.000$ & & & $.000$ & $.000$ & & \\
							   & $3$ & $.000$ & $.000$ & $.000$ & & $.000$ & $.000$ & $.000$ & \\
							   & $4$ & $.000$ & $.000$ & $.000$ & $.000$ & $.008$ & $.000$ & $\mathbf{0.836}$ & $.000$ \\

\hline
\end{tabular}
\caption{$p$-values resulting from the Kruskal-Wallis test on the audio model. Results are provided for different order of tasks.}
\label{tab:kruskal}
\end{center}
\end{table}

\subsection{Model efficiency}

We measure model efficiency using static metrics: amount of memory required to store weights and FLOPS; and dynamic metrics: inference time and energy consumption. For dynamic measurements, we measured the inference time and energy consumption of one-data-sample execution on a Google Pixel 2 Android smartphone with only CPU as the computing unit. This device is equipped with octa-core 1.9 GHz CPU and 4 GB memory. We only provide results for the audio model as it is more complex than the activity one. 

Our continual learning model uses $5.1$ MB of memory, whereas each single-task model uses between $4.9$ and $5$ MB. That is, each single task models occupies approximately the same storage as the continual model. In general, these values almost double those reported in related embedded systems research using non-bayesian approaches and similar architectures~\cite{georgiev2017low}. The reason is that our model needs to store, apart from the mean of each parameter, its variance. As for the computational cost, $252.34m$ operations are needed with the continual learning model, whereas single task models would need $250.08m$ operations for speaker recognition, $246.17m$ for ambient detection, $245,76m$ for stress detection and $246.38m$ for emotion recognition. 

We built an application for Android to obtain the dynamic metrics. It consists of a foreground service that loads the trained model for audio tasks without any further optimization, and executes inferences. We measured the time it took to perform each inference (separately) in the test set, obtaining that, on average, it takes $1.235$ seconds to execute one inference ($std=0.079$). In order to estimate energy consumed by the each inference execution, we modify the application to execute 10, 20, 30, 40 and 50 one-data-sample inferences in each run and, for each run, we obtained the battery statistics using the dumpsys shell command. We divided the estimated power used by the application in each run by the number of inferences performed, obtaining that each one-data-sample inferences uses, on average, $0.238$ $mAh$ ($std=0.001$). Using the following conversion to Joules $\frac{0.238}{1000}x3600x3.85$, this is equivalent to $3.30$ $Joules$. Given that a standard Pixel 2 battery can deliver $2700$ $mAh$, or $\frac{2700}{1000}x3600x3.85 = 37422$ $Joules$, an inference consumes on average $0.0088\%$ of the estimated device capacity.

There are at least two ways in which this could be more time efficient: either by performing state of the art optimisations for running models on resource constrained devices such as quantization or pruning, or by executing less forward passes per inference at the expenses of lowering the quality of the uncertainty estimations. For example, if instead of running $100$ forward passes we run $10$, the time to perform one inference would be reduced to $0.126$ seconds ($std=0.013$), whereas the estimated power usage would be $0.024$ $mAh$ ($std=0.001$). However, this is not the aim of this paper which simply wants to show the feasibility of this way of modelling.

\section{Conclusion}
\label{sec:conclusion}

We have presented and tested a Bayesian inference based continual learning framework for learning sensing-based analysis tasks. This framework extends the continual learning framework proposed by Nguyen \emph{et al.}~\cite{nguyen2018variational} by adding an extra hyperparameter to the loss function that \emph{regulates} the weight of the current and previous tasks in the parameters of the model. We evaluated this framework on four different audio analysis tasks namely speaker identification, ambient scene detection, stress detection and emotion recognition, and two activity-related analysis tasks namely human activity recognition and person identification. Our experiments showed that our learning method is robust against changes in the order of tasks to be learned, and that the predictive entropy and the mutual information between the prediction and the posterior over the model parameters for those samples wrongly classified are usually higher than for those correctly classified, which may be used to accept or discard predictions, especially useful in critical applications. We also demonstrated the feasibility of deploying our trained continual learning model for audio tasks in an Android smartphone as an example of the use of the framework in the wild on edge devices.
\section*{Acknowledgments}
This work was supported by ERC through Project 833296 (EAR) and by Nokia Bell Labs through their donation for the Centre of Mobile, Wearable Systems and Augmented Intelligence. 

\bibliographystyle{IEEEtran}
\bibliography{IEEEabrv,ms}

\end{document}